%% file: main.tex
\documentclass[10pt,twocolumn,letterpaper]{article}
\usepackage[inline]{enumitem}
\usepackage{color}
\usepackage{makecell}
\usepackage[inkscapelatex=false]{svg}
\usepackage{multirow}
\usepackage{amssymb}
\usepackage{pifont}
\usepackage{indentfirst}

\usepackage[pagenumbers]{cvpr} 


\definecolor{cvprblue}{rgb}{0.21,0.49,0.74}
\usepackage[pagebackref,breaklinks,colorlinks,citecolor=cvprblue]{hyperref}

\def\paperID{5391} 
\def\confName{CVPR}
\def\confYear{2024}

\title{Rotated Multi-Scale Interaction Network for Referring Remote \\ Sensing Image Segmentation}

\author{Sihan Liu\footnotemark[1],\:
Yiwei Ma\footnotemark[1],\:
Xiaoqing Zhang\footnotemark[1],\:
Haowei Wang,\:
Jiayi Ji\footnotemark[2],\:
Xiaoshuai Sun,\:
Rongrong Ji\\
Key Laboratory of Multimedia Trusted Perception and Efficient Computing, \\ Ministry of Education of China, Xiamen University, 361005, P.R. China \\
{\tt\small \{liusihan, yiweima, 36920221153149, wanghaowei\}@stu.xmu.edu.cn }\\
{\tt\small jjyxmu@gmail.com ~~~~ \{xssun, rrji\}@xmu.edu.cn}
 }

\begin{document}
\maketitle

\renewcommand{\thefootnote}{\fnsymbol{footnote}}
\footnotetext[1]{These authors contributed equally to this work.} 
\footnotetext[2]{The corresponding author.}

\input{sec/0_abstract}
\input{sec/1_intro}
\input{sec/2_related_work}
\input{sec/3_Dataset}

\input{sec/4_method}

\input{sec/5_experiment}

\input{sec/6_conclusion}

\newpage
{
    \small
    \bibliographystyle{ieeenat_fullname}
    \bibliography{main}
}


\end{document}

%% file: sec/0_abstract.tex
\begin{abstract}
Referring Remote Sensing Image Segmentation (RRSIS) is a new challenge that combines computer vision and natural language processing. 
Traditional Referring Image Segmentation (RIS) approaches have been impeded by the complex spatial scales and orientations found in aerial imagery, leading to suboptimal segmentation results. 
To address these challenges, we introduce the Rotated Multi-Scale Interaction Network (RMSIN), an innovative approach designed for the unique demands of RRSIS.
RMSIN incorporates an Intra-scale Interaction Module (IIM) to effectively address the fine-grained detail required at multiple scales and a Cross-scale Interaction Module (CIM) for integrating these details coherently across the network. 
Furthermore, RMSIN employs an Adaptive Rotated Convolution (ARC) to account for the diverse orientations of objects, a novel contribution that significantly enhances segmentation accuracy. 
To assess the efficacy of RMSIN, we have curated an expansive dataset comprising 17,402 image-caption-mask triplets, which is unparalleled in terms of scale and variety.
This dataset not only presents the model with a wide range of spatial and rotational scenarios but also establishes a stringent benchmark for the RRSIS task, ensuring a rigorous evaluation of performance.
Experimental evaluations demonstrate the exceptional performance of RMSIN, surpassing existing state-of-the-art models by a significant margin. Datasets and code are available at~\url{https://github.com/Lsan2401/RMSIN}.

\end{abstract}

%% file: sec/1_intro.tex
\section{Introduction}
\label{sec:intro}

\begin{figure}
    \centering
    \includegraphics[width=\linewidth]{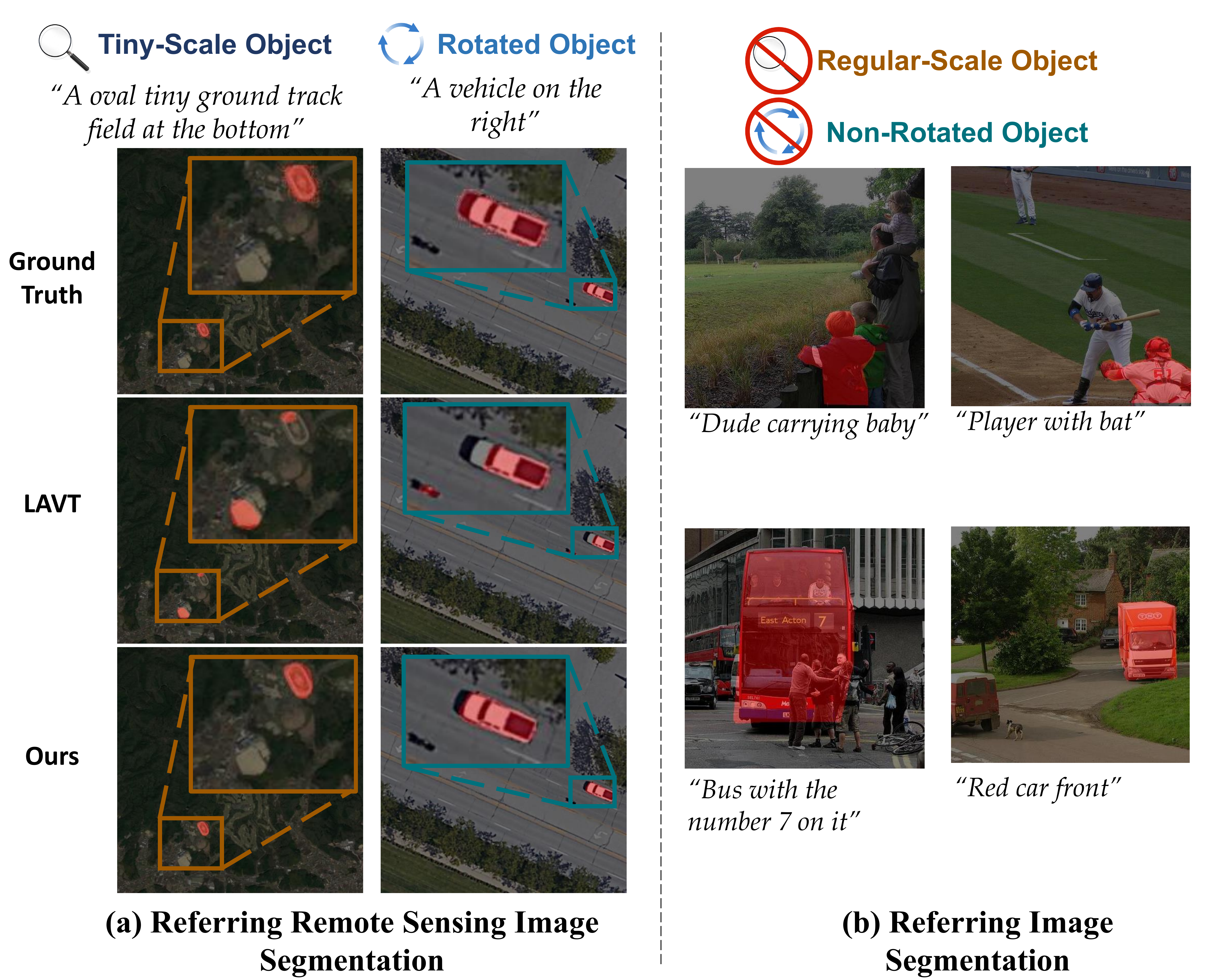}
    \vspace{-1em}
    \caption{Comparison between the newly constructed RRSIS-D and conventional RIS datasets~\cite{yu2016modeling}, highlighting the complex spatial scales and orientations prevalent in aerial imagery. (a) Examples from our RRSIS-D, demonstrating the limitations of traditional RIS methods (\emph{e.g.}, LAVT~\cite{9880242}) in handling such complexities. (b) Examples from a standard RIS dataset~\cite{yu2016modeling}.}
    \label{fig1}
    \vspace{-1em}
\end{figure}


Referring Remote Sensing Image Segmentation (RRSIS) stands at the forefront of integrating computer vision with natural language processing~\cite{ma2022xclip, ma2023towards, ji2022knowing,luo2021dual}, aiming to segment specific areas from aerial images based on textual descriptions. This sophisticated task goes beyond the capabilities of traditional RIS~\cite{hu2016segmentation, Liu_2017_ICCV, Li_2018_CVPR}, requiring an intricate understanding of the spatial and geographic nuances conveyed from aerial perspectives. 
RRSIS plays a crucial role in a wide range of applications, including land use categorization~\cite{Foody2003RemoteSO}, climate impact studies~\cite{Rolnick2019TacklingCC}, and urban infrastructure management~\cite{Duan2016TowardsLC}. By pushing the boundaries of semantic understanding in remote sensing data, RRSIS is advancing the possibilities in these domains.
Despite this, the field has been constrained by the limited scale and scope of existing datasets, which are insufficient for training models to the level of accuracy required for these critical tasks.

In light of these requirements, our research introduces an expansive new benchmark dataset, namely RRSIS-D, designed to propel the development of RRSIS. 
This dataset surpasses its predecessors\footnote{Known as RefSegRS, as of November 17, 2023, this dataset is not yet publicly available.}~\cite{yuan2023rrsis} in threefold size, encompassing not only higher resolution images but also a significantly broader range of geographic diversity.
The development of this dataset is guided by the Segment Anything Model (SAM)~\cite{kirillov2023segment}, which facilitates a semi-automated annotation process, thereby mitigating the labor-intensive nature of generating accurate pixel-level masks traditionally.
This process involves deriving initial segmentation masks from bounding box prompts and refining them to ensure high fidelity to the complex reality of aerial imagery. The result is a comprehensive corpus of 17,402 remote sensing image-caption-mask triplets, an invaluable resource aimed at advancing the precision and utility of RRSIS.

Furthermore, although the existing RIS methodologies~\cite{hui2020linguistic, Li_2018_CVPR,wang2021cris, 10132374} have demonstrated effectiveness in specific domains~\cite{yu2016modeling, 45606, nagaraja16refexp}, they face limitations when applied to the diverse and intricate nature of remote sensing imagery.
As illustrated in \cref{fig1}, aerial images pose distinct challenges that are not encountered in conventional image datasets, including vast and diverse spatial scales, as well as objects captured from multiple orientations.
Current RIS approaches typically excel in aligning visual and linguistic elements in well-bounded contexts~\cite{ye2019cross, Huang_2020_CVPR} but falter when faced with the chaotic and unstructured nature of aerial images. 
The inability of these methods to grapple with high levels of spatial variation and rotational diversity results in a notable performance gap in RRSIS tasks, highlighting the need for a more robust and versatile approach.

To overcome the inherent limitations in existing approaches, we present the Rotated Multi-Scale Interaction Network (RMSIN), a pioneering architectural solution meticulously designed to tackle the complexities of RRSIS.
Our approach introduces a sophisticated Intra-scale Interaction Module (IIM) that excels at extracting detailed features within individual layers, as well as a Cross-scale Interaction Module (CIM) that facilitates comprehensive feature fusion across the entire network.
Furthermore, we integrate an Adaptive Rotated Convolution (ARC) into the decoder, empowering the model to effectively handle the intricate rotational variations exhibited by objects.
By seamlessly integrating these modules, RMSIN proficiently extracts and aligns features across diverse scales and orientations, resulting in remarkable performance enhancements for RRSIS.

To sum up, our key contributions are as follows: 
\begin{itemize}
\item We introduce RRSIS-D, a novel benchmark dataset tailored for Referring Remote Sensing Image Segmentation (RRSIS). This dataset accommodates substantial variations in both spatial resolution and object orientation.

\item We propose the Rotated Multi-Scale Interaction Network (RMSIN) to address the challenges posed by the multiple spatial scales and orientations prevalent in aerial imagery.

\item We propose IIM and CIM to handle fine-grained information within and across different scales. Meanwhile, We design ARC to enhance the model's robustness against the ubiquitous rotational phenomena in RRSIS.

\end{itemize}


%% file: sec/2_related_work.tex
\section{Related work}
\textbf{Referring Image Detection and Segmentation. }
Referring Image Detection aims to predict a bounding box corresponding to a given referring expression. Existing works can be classified into two-staged methods~\cite{Hong2019LearningTC, Hu_2017_CVPR, Zhang_2018_CVPR, Zhuang_2018_CVPR} which are based on region proposal ranking, and one-stage methods~\cite{Cai2022XDETRAV, Li2021ReferringTA, Liao_2020_CVPR, Yang_2019_ICCV,luo2020multi,luo2022towards,luo2023towards,zhu2022seqtr} which directly predict the target bounding box. 
Referring Image Segmentation aims to achieve pixel-level localization of target objects within images based on associated referring expressions, presenting a more complex task~\cite{hu2016segmentation}. Early works~\cite{Liu_2017_ICCV, Li_2018_CVPR, nagaraja16refexp} leverage convolution networks and recurrent neural networks to extract vision and language features, respectively. These features are then fused by simple concatenation to generate final predictions.
Subsequent methods~\cite{Chen_2019_ICCV, Liu_2017_ICCV, Chen_lang2seg_2019, Jing_2021_CVPR, Margffoy-Tuay_2018_ECCV, wang2021cris, 10132374, Kamath_2021_ICCV, vision-language-transformer,wang2023towards,wu20233d,yang2023semi} mainly focus on vision-language alignment to enhance predictions. Some employ recurrent refinement~\cite{Chen_2019_ICCV, Liu_2017_ICCV}, while others utilize dynamic filters~\cite{Chen_lang2seg_2019, Jing_2021_CVPR, Margffoy-Tuay_2018_ECCV} to fuse visual and linguistic features. Recently, leveraging the Transformer's outstanding performance, methods can deviled into two categories: those performing cross-modal decoder fusion based on Transformer~\cite{wang2021cris, 10132374, Kamath_2021_ICCV, vision-language-transformer} and those incorporating language-aware visual encoding instead of post-feature fusion~\cite{9880242, Wu2022TowardsRR, Kim_2022_CVPR}. However, due to the specific characteristics of aerial images, these methods exhibit limited performance in the Remote Sensing field. Some approaches~\cite{Kim_2023_ICCV,slvit,8953566} have introduced extra scale interaction modules to enhance feature extraction. However, the extreme semantic gap between natural images and aerial images still results in suboptimal performance.

\noindent\textbf{Remote Sensing Referring  Image Detection and Segmentation.}
Referring Image Detection in Remote Sensing field is a novel task with limited research. It was first introduced by~\cite{Sun2022VisualGI}, where a new dataset and a baseline model were proposed. Recently, the transformer-based method RSVG~\cite{10056343} has been proposed. RSVG utilized the Vision Transformer~\cite{dosovitskiy2020vit} and BERT~\cite{Devlin2019BERTPO} as backbones, incorporating the Multi-Level Cross-Modal feature learning module to address multi-scale variations in aerial images.
Remote Sensing Referring Image Segmentation (RRSIS) is also a nascent field owing to the aforementioned challenges.~\citet{yuan2023rrsis} constructed the first RRSIS dataset and proposed a model that utilizes the deep and shallow feature interactions to enhance the multi-scale feature extraction. However, this model encounters limitations in handling more complex datasets. In an effort to address the existing gaps in RRSIS, we propose a more extensive and intricate dataset, RRSIS-D, alongside a novel model named RMSIN and conduct a comparative evaluation of the performance of \citet{yuan2023rrsis}'s model on our dataset.

%% file: sec/3_Dataset.tex
\section{RRSIS-D}

\begin{figure}
    \centering
    \includegraphics[width=0.9\linewidth]{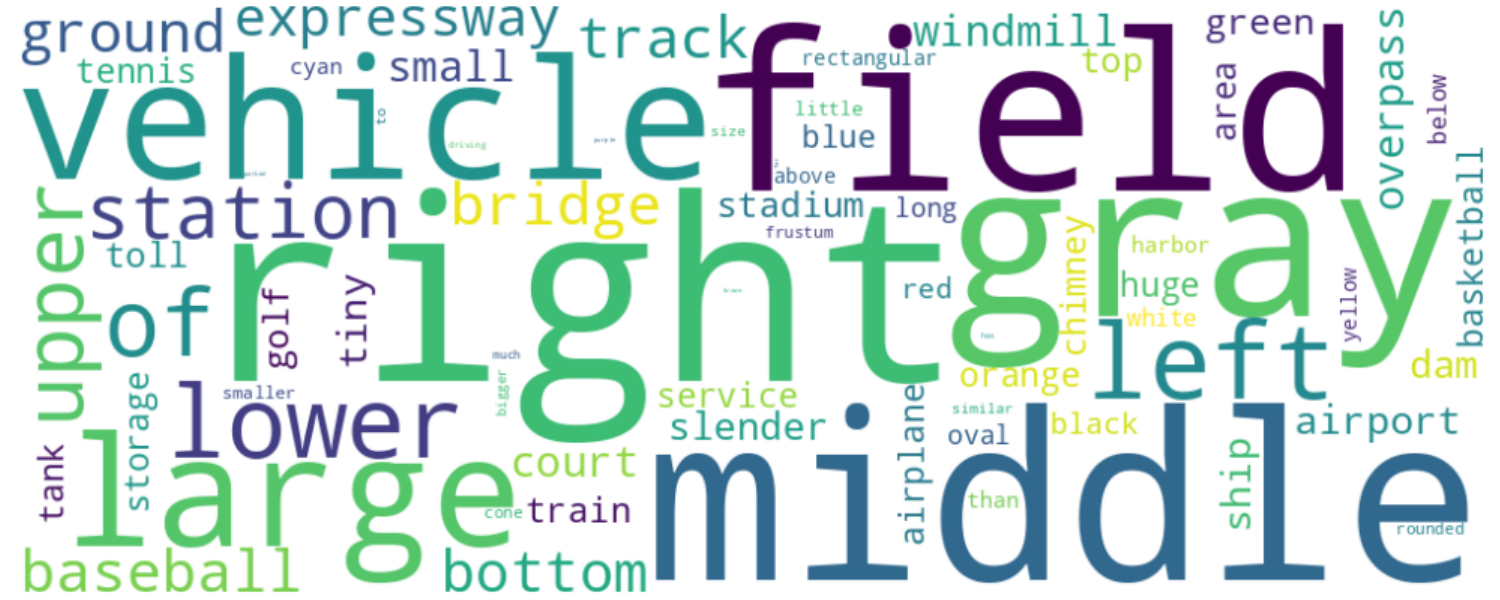}
    \caption{Word cloud for top 100 words within the expressions of RRSIS-D.}
    \label{fig:wordcloud}
\end{figure}

\begin{figure}
    \centering
    \includegraphics[width=0.82\linewidth]{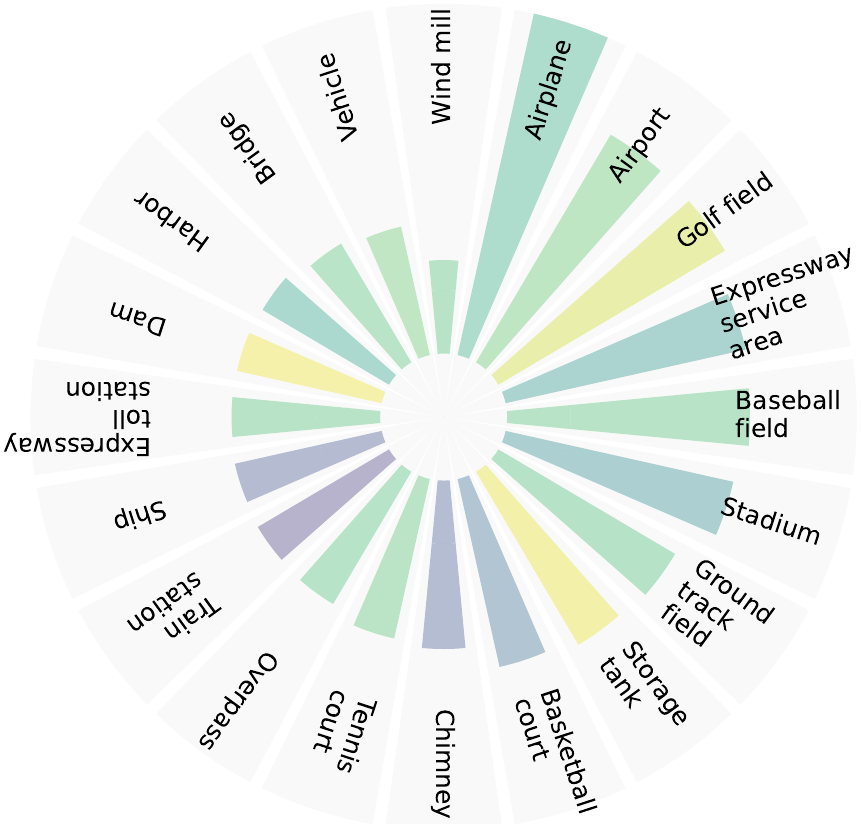}
    \caption{Distribution of image categories of RRSIS-D.}
    \label{fig:distribution}
\end{figure}

We present a new large-scale benchmark, called \emph{RRSIS-D}, specifically designed for the RRSIS task.
Fig.~\ref{fig:wordcloud} depicts the word cloud representation of this dataset.
Motivated by the exceptional segmentation performance achieved by the Segment Anything Model (SAM)~\cite{kirillov2023segment}, we adopt a semi-automatic approach that capitalizes on bounding boxes and SAM to generate pixel-level masks, resulting in cost savings during the annotation process.
Specifically, we follow the steps outlined below to generate pixel-wise annotations for language expressions:
\begin{itemize}

\item  \textit{Step 1}. Pixel-level masks for all images in the dataset are generated by leveraging the bounding box prompts provided by the RSVGD Dataset~\cite{10056343} through the employment of SAM. It is noteworthy, however, that the performance of SAM may exhibit variability in accuracy when applied to partial images, owing to the inherent domain gap between aerial and natural images.

\item  \textit{Step 2}. We undertake a manual refinement process for masks associated with problematic aerial images. This refinement involves the utilization of a filling algorithm to address hollow problems within the masks. Subsequently, a meticulous curation of the dataset is conducted to identify problematic data, and manual annotation is employed to generate masks aligned with annotation standards. This manual annotation process is facilitated by the software tool~\cite{ISAT_with_segment_anything} designed in accordance with the principles of SAM, ensuring the accurate generation of masks corresponding to linguistic expressions.

\item  \textit{Step 3}. To enhance the compatibility of RRSIS-D with natural RIS models, finally, the annotations are converted into RefCOCO dataset~\cite{lin2014microsoft} format for better usability.
\end{itemize}

\begin{figure}
	\centering
       \includegraphics[width=\linewidth]{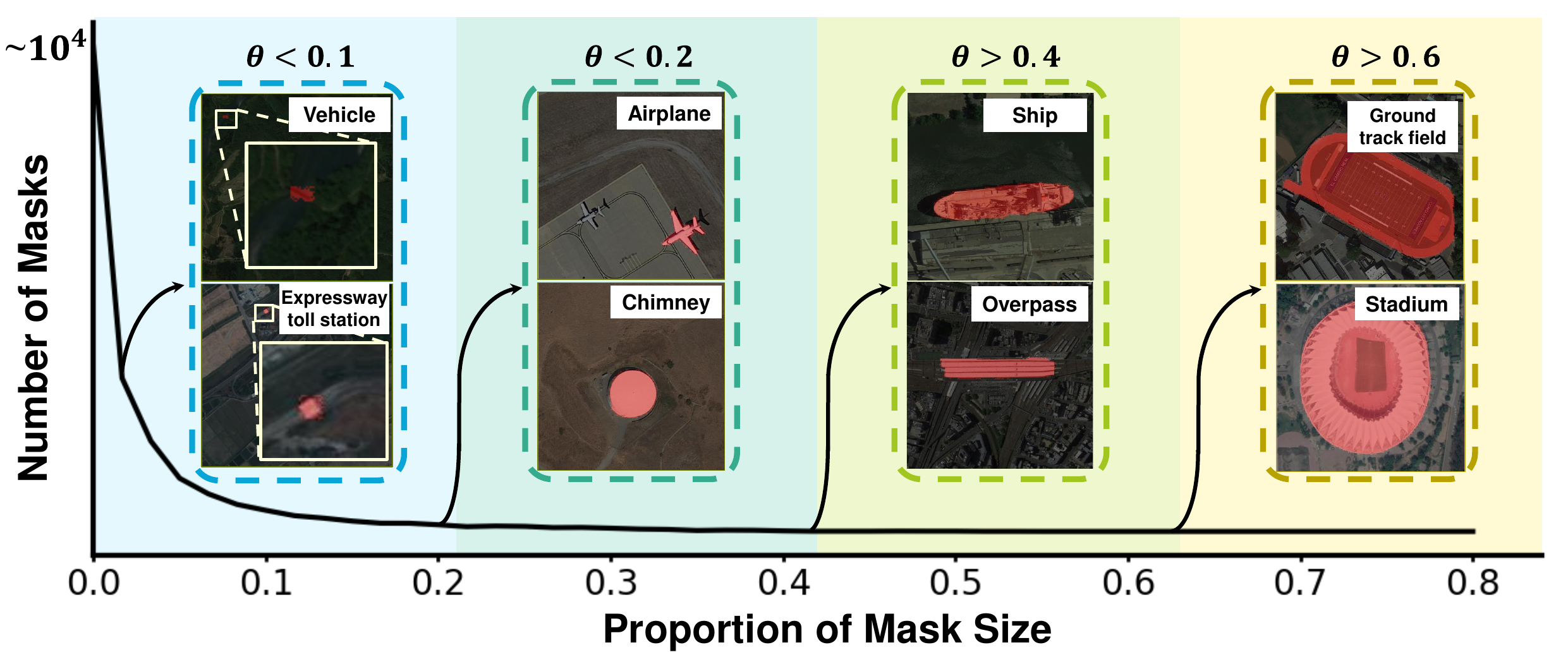}
	\caption{Distribution of mask sizes, with the horizontal axis showing mask coverage percentage in images ($\theta$) and the vertical axis representing total mask count, illustrated with varied-size ground truth examples.}
    \label{fig:mask}
    \vspace{-1em}
\end{figure}

The benchmark statistics, as presented in \cref{tab:example}, exhibit notable distinctions from the existing RefSegRS dataset~\cite{yuan2023rrsis}.
Our proposed dataset, RRSIS-D, comprises a comprehensive collection of 17,402 images, accompanied by their corresponding masks and referring expressions.
A standardized resolution of 800px in height and 800px in width has been uniformly applied to all images. 
Furthermore, the semantic labels comprise 20 categories, supplemented by 7 attributes, thereby enhancing the semantic richness of the referring expressions.
To illustrate the prevalence of each category, the category distribution is graphically represented in \cref{fig:distribution}.
For instance, the category ``Airplane" accounts for 15.6\% of the total, ranking highest in terms of quantity.
It is worth noting that our dataset offers enhanced flexibility in terms of mask resolution, surpassing that of RefSegRS.

The statistics of the generated masks are depicted in \cref{fig:mask}.
Notably,  a significant portion of the targets is extremely small, occupying only a fraction of the overall image. However, there are also instances of large-scale objects exceeding 400,000 pixels in size. Some examples of masks with different sizes are illustrated in the figure, highlighting the substantial variability in scale across different categories in the dataset. This presents a challenging task, as it involves predicting images with significant large-scale variations and numerous small targets.

\begin{table}
  \centering
  \renewcommand{\arraystretch}{2}
  \resizebox{\linewidth}{!}{
  \begin{tabular}{@{}l|c|c|c|c|c}
    \toprule
    Dataset & \makecell{Number\\of images} & \makecell{Image \\size} & \makecell{Spatial \\resolution} & \makecell{Attributes \\of expression} & \makecell{Mask \\generation}\\
    \midrule
    RefSegRS~\cite{yuan2023rrsis}& 4420 & 512 $\times$ 512 & 0.13m & 3 & Manually\\
    \hline
    RRSIS-D & 17402 & 800 $\times$ 800 & 0.5m $\sim$ 30m & 7 & \makecell{Semi\\automatically}\\
    \bottomrule
  \end{tabular}}
  \caption{Compare our dataset with the previous dataset. }
  \label{tab:example}
  \vspace{-1em}
\end{table}

%% file: sec/4_method.tex
\section{RMSIN}
\label{sec:method}
\subsection{Overview}
\begin{figure*}
  \centering
  \includegraphics[width=0.96\textwidth]{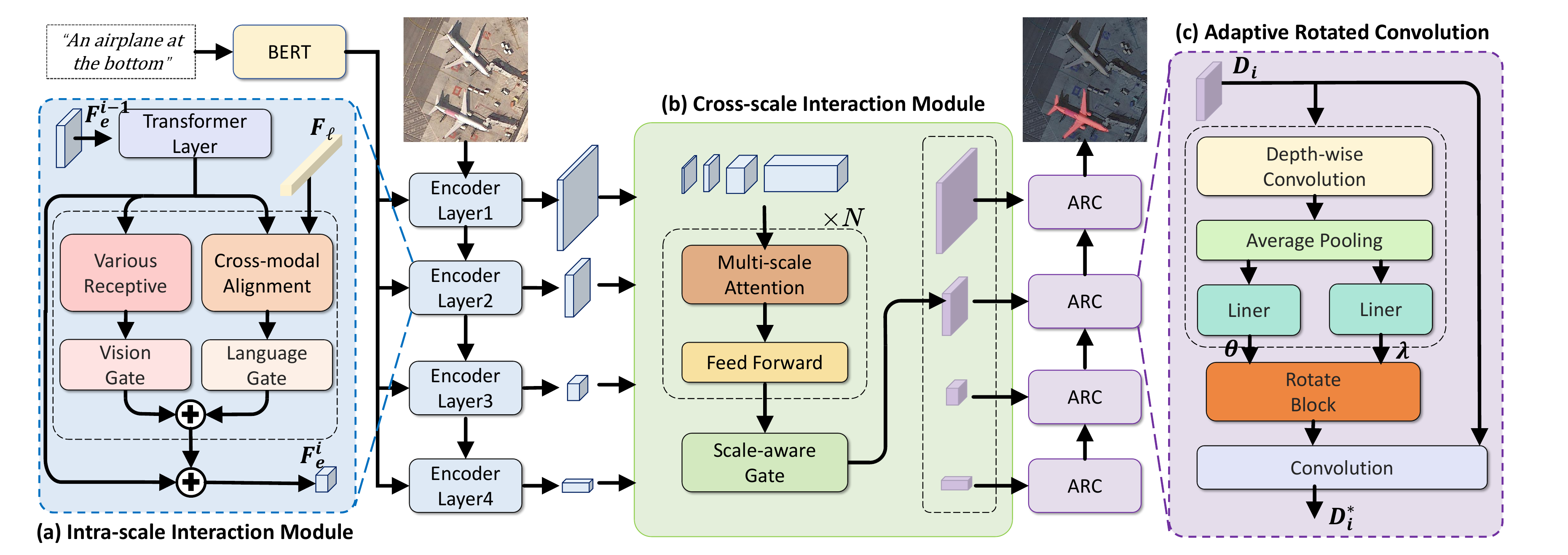}
   \caption{Overview of the proposed RMSIN.}
   \label{fig:overview}
   \vspace{-1em}
\end{figure*}

The pipeline of our proposed model is depicted in~\cref{fig:overview}.
Initially, given the input image $I \in \mathbb{R}^{H\times W\times 3}$ and the language expression $E = \{\omega_i\}, {i \in \{0, \dots, N\}}$, where $H$ and $W$ represent the height and width of the input image, and $N$ is the length of the expression, the input language expression $E$ is transformed into the feature space $F_\ell \in \mathbb{R}^{N \times C}$ via the backbone $f_\ell$.
The following Compounded Scale Interaction Encoder (CSIE), which is composed of an Intra-scale Interaction Module (IIM) at each stage, and a Cross-scale Interaction Module (CIM), is applied to generate the fused features with sufficient semantics across multiple scales.
Finally, we propose an Adaptive Rotated Convolution (ARC) based Oriented-Aware Decoder (OAD) to generate the segmentation mask by the parallel inference on the features from the multiple stages of the CSIE.

\subsection{Compounded Scale Interaction Encoder}
To effectively locate diverse targets with the guidance of the referred texts, the information for multi-scale is just as important as the referring expressions. Given the language features $F_\ell$ and the input image $I \in \mathbb{R}^{H\times W\times 3}$, the Compounded Scale Interaction Encoder (CSIE) brings about the fusion across vision-language modality in a multi-stage way with both intra- and inter-perspective. 

Specifically, the CSIE is constructed with two components, Intra-scale Interaction Module (IIM) and Cross-scale Interaction Module (CIM).


\subsubsection{Intra-scale Interaction Module}
\label{isim}
The first part of each stage in CSIE, the Intra-scale Interaction Module (IIM) is designed to further excavate the rich information within each scale and facilitate interaction between the vision and language modalities.
Based on a hierarchy of four stages, IIM could be denoted as $\{\phi_i\}_{i \in \{1,2,3,4\}}$. After obtaining the language features $F_\ell \in \mathbb{R}^{N \times C}$ through the text backbone, where $C$ denotes the number of channels, the output features $F_{e}^i$ of IIM at stage $i$ could be described as:
\begin{equation}
  F_{e}^i = \phi_i(F_{e}^{i-1}, F_\ell),
\end{equation}
where $F_{e}^0$ is extracted from the vision backbone $f_v$ with the input $I$. More detailed, during the stage $i$, the input features $F_{e}^{i-1}$ undergo a combination of downsampling and MLP~\cite{liu2021swin} to reduce the scale and unify the dimension of features, resulting in $\hat{F}_{e}^{i-1}$. 
The downsampled input $\hat{F}_{e}^{i-1}$ is fed into two branches for enhancing visual priors and fusing cross-modal information individually.

\textbf{Various Receptive Branch} is the first branch. The feature $\hat{F}_{e}^{i-1}$ is transformed through multiple branches with different settings of convolution kernels to yield features with various receptive fields, which could be formulated as:
\begin{equation}
    \omega^i = \sigma \left(\sum_{j=0}^{J}\left(\frac{1}{C}\sum^C k_j^i \ast \hat{F}_{e}^{i-1}\right)\right),
\end{equation}
where the $k_j^i$ means the $j$-th branch of convolution and the $\sigma$ is the Sigmoid Function. The above formulation indicates that the different convolution setting is utilized to balance the weight $\omega^i \in (0, 1)^{H \times W}$ between all the pixels. The weight is taken to enhance the features by:
\begin{equation}
    \hat{F}_{e1}^{i-1} = \omega^i \otimes \hat{F}_{e}^{i-1}.
\end{equation}

In addition, the output is regulated by a Visual Gate, adding to the raw image features as a complement to local detail information. The specific implementation of the gate is:
\begin{equation}
 \alpha = {\rm Tanh(Linear(ReLU(Linear(}\hat{F}_{e1}^{i-1})))),
 \label{eq4}
\end{equation}
where ${\rm Linear(\cdot)}$ denotes the linear projection, and ${\rm Tanh}(
\cdot)$ and ${\rm ReLU(\cdot)}$ represent the activation functions.
 

\textbf{Cross-modal Alignment Branch} is designed for multi-modal alignment, which is the key to enabling the model to comprehend natural language. 

Concretely, taking the input $\hat{F}_{e}^{i-1}$ and language features $F_\ell$, the module first implements scaled dot-product attention \cite{Vaswani2017AttentionIA} using $\hat{F}_{e}^{i-1}$ as the query and $F_\ell$ as the key and value to obtain the multi-modal features:
\begin{equation}
  A^i = {\rm attention}(\hat{F}_{e}^{i-1}W^i_q, F_\ell W^i_k, F_\ell W^i_v),
\end{equation}
where $W^i_q$, $W^i_k$ and $W^i_v$ are the linear projection matrices.
Subsequently, the attention $A^i$ is combined with $\hat{F}_{e}^{i-1}$ to obtain language-guided image features:
\begin{equation}
 \hat{F}_{e2}^{i-1} = {\rm Proj}(A^iW^i_w \otimes \hat{F}_{e}^{i-1}W^i_m),
\end{equation}
where $W^i_w$ and $W^i_m$ are the projection matrices, and $\otimes$ denotes element-wise multiplication. The obtained result is passed through a final $1 \times 1$ convolution, denoted as ${\rm Proj}(\cdot)$, to produce the final output.

Similar to the operation performed on the output of $\hat{F}_{e1}^{i-1}$, the result is regulated by $\beta$ from the Linguistic Gate shares an identical structure with the Visual Gate and is added to the raw image features, serving as supplementary linguistic features.
Consequently, the overall output features of IIM at stage $i$ can be illustrated as:
\begin{equation}
    F_{e}^{i} = \hat{F}_{e}^{i-1} + \alpha\hat{F}_{e1}^{i-1} + \beta\hat{F}_{e2}^{i-1}.
 \label{eq7}
\end{equation}


\subsubsection{Cross-scale Interaction Module}
\label{csim}
While the IIM adequately extracts localized multi-scale information guided by linguistic features, we additionally design a Cross-scale Interaction Module (CIM) to further enhance the interaction between the coarse and fine stages, particularly in response to the scale variation challenge observed in aerial images. Specifically, the module takes features collected from each stage of the IIM, {\emph{i.e}\onedot}, the previously mentioned $F_e^i,\ i \in \{1, 2, 3, 4\}$ as input and performs multi-stage interaction. The structure is depicted schematically in~\cref{fig:overview}.

\textbf{Multi-stage Feature Combination} is first performed, where the features ${F_e^i}$ are downsampled to the same size and concatenated along the channel dimension. The formula expression is as follows: 
\begin{equation}
\begin{aligned}
  &F_d^i = {\rm downsample}({F_e^i}),\quad i \in \{1, 2, 3, 4\},\\
  &F_c^\ast= \mathop{\rm concat}\limits_{c}(F_d^1, F_d^2, F_d^3, F_e^4),\\
\end{aligned}    
\end{equation}
where $F_d^i$ represents the downsampled features, and $F_c^\ast$ represents the multi-stage feature concatenated along the channel dimension. ${\rm downsample}(\cdot)$ is typically implemented through average pooling. 

\textbf{Multi-scale Attention Layer} is subsequently implemented. Specifically, we design various perceptive fields for the concatenated feature $F_c^\ast$ to achieve deep multi-scale interaction. $F_c^\ast$ is resized to different scales through the depth-wise convolutions with diverse kernel sizes and strides, defined as follows:
\begin{equation}
\begin{aligned}
  &F_c^m = k^m \ast F_c^\ast, \\
  &h^m = \lfloor {\frac{h - 1}{m} + 1} \rfloor, w^m =  \lfloor {\frac{w - 1}{m} + 1} \rfloor,
\end{aligned}    
\end{equation}
where $m \in \{1,\dots,M\}$, $M$ is the number of resized scales, $k_m$ is the $m$-th depth-wise convolution and $h_m$ and $w_m$ are the corresponding height and weight of the $F_c^m$.
With the set $\{F_c^m|m \in \{1,\dots,M\}\}$, we flatten all the elements on the size dimension and concatenate them as a sequence features $\hat{F}_c^\ast \in \mathbb{R}^{(\sum_{1}^{M} h^m \times w^m) \times C}$. Similar to vanilla attention~\cite{Vaswani2017AttentionIA}, we take the origin feature $F_c^\ast$ as the query, and the multi-scale-aware feature $\hat{F}_c^\ast$ as the key and value to perform cross-scale interaction:

\begin{equation}
    \widetilde{F}_c^\ast = {\rm softmax}(\frac{F_c^\ast W_q \cdot \hat{F}_c^\ast W_k^T}{\sqrt{C}}) \cdot \hat{F}_c^\ast W_v.
\end{equation}
For better preservation of local details, following inspiration from HRViT~\cite{Gu_2022_CVPR}, a local relationship compensation called LRC is incorporated to regulate the output of the multi-scale attention. Consequently, the final output of the Multi-scale Attention Layer is expressed as:
\begin{equation}
    F_c = \widetilde{F}_c^\ast + {\rm DWConv(Hardswish}(F_c^\ast)),
\end{equation}
where ${\rm DWConv}(\cdot)$ represents depth-wise convolution, and ${\rm Hardswish}(\cdot)$ is the activity function, implemented in accordance with~\cite{Gu_2022_CVPR} to enhance the extraction of multi-scale local information.

The Feed Forward Layer follows the Multi-scale Attention layer which is identical to the standard attention block~\cite{Vaswani2017AttentionIA}. The feature $F_c$ is divided into four parts to revert to the size of $F_e^i$ by upsampling and subsequently fed into the Scale-aware Gate to obtain the final output.

\textbf{Scale-aware Gate} is employed to alleviate the semantic gap before and after multi-scale attention. Specifically, for each part from $F_c$, we implement the corresponding part from $F_e$ to measure the weight of the cross-scale interaction. This weight is considered as the assistance residual for the features from IIM. The formulation is as follows:
\begin{equation}
    F_o^i = {\rm sigmoid}(F_e^i W_1) \otimes F_c^i W_2 + F_e^i W_3,
\end{equation}
where $i \in \{1, 2, 3, 4\}$. The output of the Scale-aware Gate is utilized in the decoder for final mask prediction.


\subsection{Oriented-aware Decoder}
\label{oad}
The set of features $\{F_o^i|i \in \{1, 2, 3, 4\}\}$ from the CSIE are used to generate the segmentation. Considering that object instances in aerial images often exhibit various orientations, using static horizontal convolution kernels for mask generation may result in a loss of precision. Inspired by oriented object detection, where the problem has been researched for decades and achieved considerable progress~\cite{9001201, Pu_2023_ICCV, Yang2021LearningHB, Yang2022TheKL}, we incorporate the Adaptive Rotated Convolution (ARC) into the segmentation decoder tailored for the specific needs of RRSIS task to achieve better mask prediction.

\subsubsection{Adaptive Rotated Convolution}
The ARC captures angle information from input features and dynamically re-parameterizes the kernel weights to filter out redundant features.
Specifically, it extracts orientation features and predicts n angles $\theta \in \{1, \dots, n\}$ and corresponding weights $\lambda \in \{1, \dots, n\}$ based on the input. For the input $X$, the $\theta$ and $\lambda$ are predicted as:
\begin{equation}
    \theta,\ \lambda  = {\rm Routing}(X),
\end{equation}
where the concrete structure of the Routing Block is illustrated in~\cref{fig:overview}. The static convolution kernel weights can be viewed as specific sampling points from the two-dimensional kernel space. Thus, the rotation of the convolution kernel is the process of rotary resampling. 
Specifically, the convolution kernel weights $W_i$ are re-parameterized according to the predicted angels as follows:
\begin{equation}
\begin{aligned}
    &Y_i^{'} = M^{-1}(\theta_i)Y_i,\\
    &W_i^{'} = {\rm interpolation}(W_i, Y_i^{'}),
\end{aligned}
\end{equation}
where $Y_i$ is the coordinates of original sampling points, $ M^{-1}(\theta_i)$ is the inverse matrix of the rotation matrix for affine transformation by angle $\theta$ around the origin, and $\rm interpolation(\cdot)$ is implemented as bilinear interpolation. 
Finally, the features are filtered by the obtained convolution kernel and subjected to a weighted sum operation to produce orientation-aware features:
\begin{equation}
   X^* = X \ast \sum_{i=1}^{n}\lambda_i W_i^{'}. 
\end{equation}
The overall top-down process of mask prediction can be concluded as follows:
\begin{equation}
\begin{aligned}
    &D_4 = F_o^4,\\
    &D_i = {\rm Seg(ARC}([D_{i+1};\ F_o^i ])),\quad i \in \{1, 2, 3\},\\
    &D_0 = {\rm Proj}(D_1),
\end{aligned}
\end{equation}
where ${\rm Seg}(\cdot)$ refers to a nonlinear block comprising a $3 \times 3$ convolution layer, a batch normalization layer, and a ReLU activity function to enhance the nonlinearity of the segmentation feature space. And ${\rm Proj}(\cdot)$ is implemented as a linear projection function to map the final feature $D_1$ into two class scores. It is notable that half of the convolution layers are replaced by the ARC to leverage orientation information in the feature space, thereby eliminating redundancy for enhanced accuracy in boundary details.

%% file: sec/5_experiment.tex
\section{Experiments}
\label{sec:exp}

\begin{table*}
  \centering
  \resizebox{\textwidth}{!}{
  \begin{tabular}{@{}p{2.3cm}|>{\centering\arraybackslash}p{1.1cm}|>{\centering\arraybackslash}p{1.1cm}|>{\centering\arraybackslash}p{0.7cm}|>{\centering\arraybackslash}p{0.7cm}
  |>{\centering\arraybackslash}p{0.7cm}|>{\centering\arraybackslash}p{0.7cm}|>{\centering\arraybackslash}p{0.7cm}|>{\centering\arraybackslash}p{0.7cm}|>{\centering\arraybackslash}p{0.7cm}|>{\centering\arraybackslash}p{0.7cm}|>{\centering\arraybackslash}p{0.7cm}|>{\centering\arraybackslash}p{0.7cm}|>{\centering\arraybackslash}p{0.7cm}|>{\centering\arraybackslash}p{0.7cm}|>{\centering\arraybackslash}p{0.7cm}|>{\centering\arraybackslash}p{0.7cm}}
    \toprule
    \multirow{2}{*}{Method} & \multirow{2}{*}{\makecell[c]{Visual \\ Encoder}}& \multirow{2}{*}{\makecell[c]{Text\\Encoder}}& \multicolumn{2}{c|}{P@0.5} & \multicolumn{2}{c|}{P@0.6} & \multicolumn{2}{c|}{P@0.7} & \multicolumn{2}{c|}{P@0.8} & \multicolumn{2}{c|}{P@0.9} & \multicolumn{2}{c|}{oIoU} & \multicolumn{2}{c}{mIoU} \\
    \cline{4-17}
     &&& \small Val &\small Test & \small Val &  \small Test & \small Val &  \small Test & \small Val &  \small Test & \small Val &  \small Test & \small Val &  \small Test & \small Val &  \small Test \\
    \midrule
    RRN~\cite{Li_2018_CVPR} & R-101
    & LSTM
    & 51.09 &51.07& 42.47 & 42.11& 33.04 & 32.77& 20.80 & 21.57& 6.14 & 6.37& 66.53 & 66.43& 46.06 & 45.64\\
    CSMA~\cite{ye2019cross} & R-101 & None&  55.68 & 55.32& 48.04 & 46.45& 38.27 & 37.43& 26.55 & 25.39& 9.02 & 8.15& 69.68 & 69.39& 48.85& 48.54\\
    LSCM~\cite{hui2020linguistic} & R-101 & LSTM & 57.12 & 56.02& 48.04 & 46.25& 37.87 & 37.70& 26.37 & 25.28& 7.93 & 8.27& 69.28 & 69.05& 50.36& 49.92\\
    CMPC~\cite{Huang_2020_CVPR} & R-101& LSTM & 57.93 & 55.83& 48.85 & 47.40& 38.50 & 36.94& 25.28 & 25.45& 9.31 & 9.19& 70.15 & 69.22& 50.41 & 49.24\\
    BRINet \cite{Hu_2020_CVPR} & R-101& LSTM & 58.79 & 56.90& 49.54 & 48.77& 39.65 &39.12& 28.21 & 27.03& 9.19 & 8.73& 70.73 &69.88& 51.14 &49.65\\
    CMPC$+$~\cite{Liu2021CrossModalPC} & R-101& LSTM & 59.19 & 57.65& 49.36 & 47.51& 38.67 &36.97& 25.91 & 24.33& 8.16 & 7.78& 70.14 &68.64& 51.41 &50.24\\
    LGCE~\cite{yuan2023rrsis} & Swin-B
    & BERT
    & 68.10 & 67.65& 60.52 & 61.53& 52.24 & 51.45& 42.24 & 39.62& 23.85 & 23.33& 76.68 & 76.34& 60.16 & 59.37\\
    LAVT~\cite{9880242} & Swin-B & BERT& 69.54 & 69.52& 63.51 & 63.63& 53.16 & 53.29& 43.97 & 41.60& 24.25 & \textbf{24.94}& 77.59 & 77.19& 61.46 & 61.04\\
    \hline
    \textbf{RMSIN (Ours)} & Swin-B & BERT& \textbf{74.66} & \textbf{74.26}& \textbf{68.22} & \textbf{67.25}& \textbf{57.41} & \textbf{55.93}& \textbf{45.29} &\textbf{42.55}& \textbf{24.43} & 24.53& \textbf{78.27} & \textbf{77.79}& \textbf{65.10} &\textbf{64.20}\\
    \bottomrule
  \end{tabular}
  }
  \caption{Comparison with state-of-the-art methods on the proposed RRSIS-D dataset. R-101 and Swin-B represent ResNet-101~\cite{he2016deep} and base Swin Transfomer~\cite{liu2021swin} models, respectively. The best result is bold. }
  \label{tab:overall}
  \vspace{-1em}
\end{table*}

\subsection{Implementation Details}

\noindent\textbf{Experiment Settings.}
In our experiments, the visual backbone utilizes Swin Transformer~\cite{liu2021swin}, pre-trained on ImageNet22K~\cite{5206848}, while the language backbone employs the base BERT model from HuggingFace’s library~\cite{Wolf2019TransformersSN}. The model is trained for 40 epochs using AdamW~\cite{Loshchilov2017DecoupledWD} with a weight decay of 0.01 and a starting learning rate of 3e-5, reducing according to polynomial decay.  The setup ran on four RTX 2080 GPUs with a batch size of 8.

\noindent\textbf{Metrics.}
We utilize Overall Intersection-over-Union (oIoU), Mean Intersection-over-Union (mIoU), and Precision@X (P@X) as evaluation metrics, similar to prior studies~\cite{10056343, Wu_2020_CVPR}.

\subsection{Comparison with state-of-the-art RIS methods}
In our experiments, we compared RMSIN's performance with existing state-of-the-art referring image segmentation methods on the validation and test subsets of our RRSIS-D dataset (see \cref{tab:overall}). For a fair comparison, we adopted the original implementation details of these competing methods.
Notably, RMSIN outperforms its counterparts across almost all metrics on both subsets, marking a significant improvement with a 3.64\% and 3.16\% increase in mIoU on the validation and test subsets respectively over the closest competitor, LAVT. 
This leap in performance is particularly evident in complex scenarios, such as detecting small or rotated objects, where it secured over 3.0\% gains in Precision at IoU thresholds of 0.5, 0.6, and 0.7.

\begin{figure*}
  \centering
  \includegraphics[width=1\textwidth]{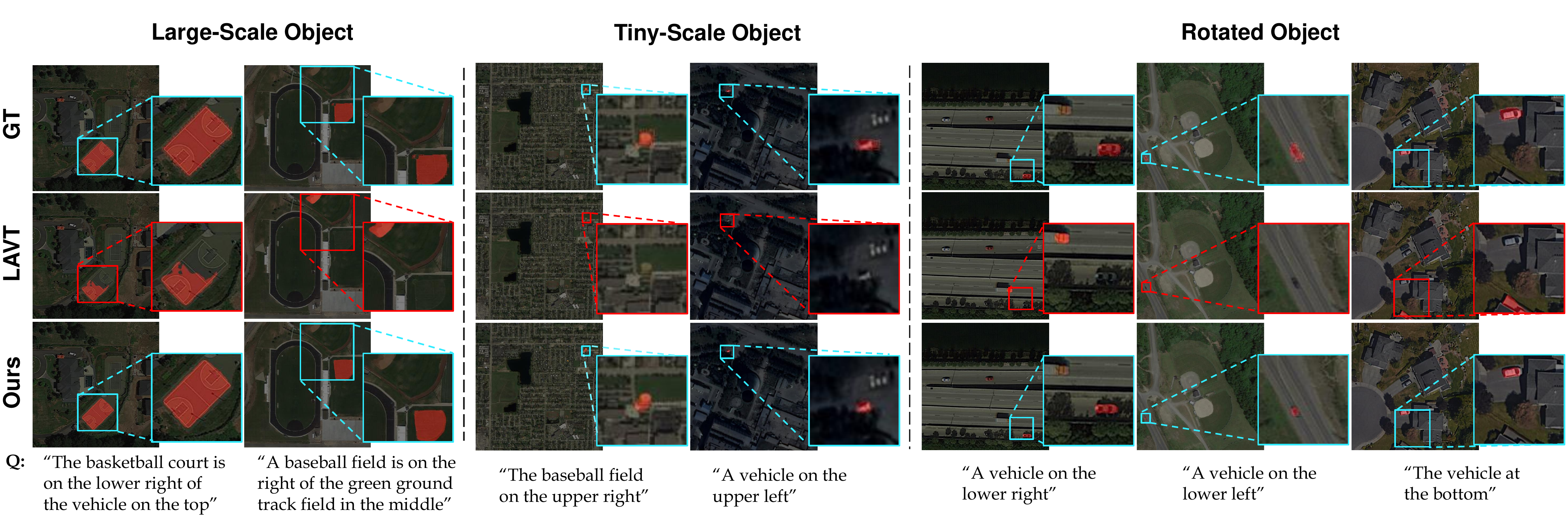}
   \caption{Qualitative comparisons between RMSIN and the previous SOTA LAVT. The left part illustrates the predictions of large-scale objects, while the middle part offers exceedingly diminutive objects amidst a highly noisy background. The right part exhibits the predictions for scenarios wherein objects are situated at diverse angles.}
   \label{fig:scale}
   \vspace{-1em}
\end{figure*}

\subsection{Ablation study}
We have performed various ablation experiments on the validation subset of RRSIS-D to assess the efficacy of the pivotal components within our proposed network.

\noindent\textbf{Effictiveness of IIM and CIM. }
To validate the efficacy of our proposed two-scale interaction modules in CSIE, we conduct ablation studies on all the combinations of IIM and CIM.
As illustrated in \cref{tab:scale}, The introduction of the IIM brings about discernible improvements in precision at lower IoU thresholds, while the incorporation of the CIM further refines predictions across various IoU levels. The combined effect of both modules demonstrates a synergistic enhancement, yielding the highest performance across all evaluated metrics, particularly in P@0.5, P@0.7, and mIoU, with margins ranging from 3.5\% to 4.5\%. These findings affirm the pivotal role played by the IIM and CIM in capturing multi-scale features from images, thus substantiating their efficacy in advancing the overall segmentation capabilities.

\begin{table}
  \centering
  \resizebox{\linewidth}{!}{
  \begin{tabular}{cc|ccccc}
    \toprule
    IIM & CIM & P@0.5 & P@0.7 & P@0.9 & oIoU & mIoU \\
    \midrule
   \ding{55}  &  \ding{55} &  69.54 & 53.16 & 24.25 & 77.59 & 61.46\\
     \ding{51} & \ding{55}  & 71.09 & 53.45 & 24.71 & 77.68 & 62.27\\
    \ding{55}  & \ding{51} &  73.68 & 56.67 & 25.69 & 77.40 & 64.25\\     
     \ding{51} & \ding{51} & \textbf{74.14} & \textbf{57.59} & \textbf{25.69} & \textbf{77.91} & \textbf{64.91}\\
    \bottomrule
  \end{tabular}
  }
  \caption{Ablation on the scale interaction modules IIM and CIM.}
  \label{tab:scale}
\end{table}

\noindent\textbf{Design options of CIM. } 
To further substantiate the effectiveness of CIM, we conduct a detailed analysis of its main components, as outlined in \cref{tab:cross-stage}. The most substantial enhancement in results is observed upon the inclusion of the complete module, showcasing the highest metric enhancement of over 4.14\%. This confirms the role of CIM in preserving local details and extracting multi-scale information. 

\begin{table}
    \centering
     \resizebox{\linewidth}{!}{
    \begin{tabular}{l|ccccc}
    \toprule
    Options &  P@0.5 & P@0.7 & P@0.9 & oIoU & mIoU \\
    \midrule
    Default & 69.54 & 53.16 & 24.25 & 77.59 & 61.46\\
    \midrule
    $+$ Multi-scale Attention& 68.91 & 53.68 & 25.11 & 77.61 & 61.46 \\
    $+$ Feed Forward & 69.83 & 52.70 &25.57 &\textbf{77.85} & 61.46\\
    $+$ LRC &\textbf{73.68} &\textbf{56.67} &\textbf{25.69} &77.40 &\textbf{64.25}\\
    \bottomrule
    \end{tabular}
    }
    \caption{Ablation on options design of CIM. Default means the vanilla self attention and we reintroduce all the designs cumulatively to demonstrate the effectiveness of each major component. }
    \label{tab:cross-stage}
\end{table}

\noindent\textbf{Design options of Decoder. } 
We explore the design of the segmentation decoder structure as demonstrated in \cref{tab: decoder}. 
The CIM yields output features with robust semantics and intricate spatial details. Thus our proposed Oriented-aware Decoder straightforwardly concatenates the features and extracts angular information through ARC to obtain more accurate predictions better suited to RS tasks. 
We also experiment with two alternative decoder structures. The exceptional results of our proposed decoder, surpassing others across all metrics, underscore the significance of incorporating angle information in the decoding process. This outcome firmly reaffirms the efficacy of our approach in customizing mask predictions for remote sensing applications, where the inclusion of precise angular information emerges as a critical factor for optimizing segmentation accuracy.

\noindent\textbf{Design options of ARC. }
We further investigate the impact of the Adaptive Rotated Convolution (ARC) replacement strategy on the results, as demonstrated in \cref{tab:oriented-decoder} (a). 
We progressively replace the convolution layers in each stage of the decoder, and the result exhibits a consistent upward trend. Consequently, we opt to replace all three layers of the decoder.
Additionally, we explore the influence of varying the number of prediction angles for ARC on the prediction results illustrated in \cref{tab:oriented-decoder} (b). 
The decoder showcases a consistent improvement in performance with an increase in the predicted number of angles, resulting in a performance boost of approximately 1\% when seated to 4 compared to 1.

\begin{table}
  \centering
  \resizebox{0.9\linewidth}{!}{
  \begin{tabular}{l|ccccccc}
    \toprule
     Decoder Design & P@0.5 &P@0.7 &P@0.9 & oIoU & mIoU \\
    \midrule
    Sum & 72.36 & 55.23 & 24.89 & 76.11 & 62.76\\
    Concat & 70.34 & 53.33 & 24.66 & 77.74 & 61.56\\
    Oriented-aware & \textbf{73.85}  & \textbf{56.84} & \textbf{25.40} & \textbf{77.76} & \textbf{64.15}\\    
    \bottomrule
  \end{tabular}
  }
  \caption{Ablation study examining decoder designs. ``Sum '' decoder employs summation instead of concatenation for cross-stage feature connection, while ``Concat '' decoder substitutes ARC in ``Oriented-aware '' decoder with static convolutions.}
  \label{tab: decoder}
\end{table}

\begin{table}
  \centering
  \resizebox{0.9\linewidth}{!}{
  \begin{tabular*}{\linewidth}{l|ccccc}
   
    \toprule
    Options &  P@0.5 & P@0.7 & P@0.9 & oIoU & mIoU \\
    \midrule
    \multicolumn{6}{l}{\footnotesize (a) Replacement of ARC in the Oriented-aware Decoder}\\
    \hline
    L = 1 & 71.72 & 54.94 & 24.43 & 75.93 & 62.53\\
    L = 2 & 72.18 & 55.29 & 24.66 & 76.37 & 62.48\\
    L = 3 & \textbf{72.36} & \textbf{55.98} & \textbf{25.00} & \textbf{77.06} & \textbf{63.81}\\
    \hline
    \multicolumn{6}{l}{\footnotesize (b) Predicted number of angles in  ARC}\\
    \hline
    n = 1& 72.36& 55.98& 25.00& 77.06& 63.81\\
    n = 2& 72.24& 54.48& 24.02& 77.39& 63.14\\
    n = 4& \textbf{73.85}& \textbf{56.84}& \textbf{25.40}& \textbf{77.76}& \textbf{64.15}\\
    \bottomrule
  \end{tabular*}
   }
  \caption{Ablation studies of ARC. L=1 indicates the replacement of the first layer in the decoder with Adaptive Rotated Convolution. Experiments on the predicted number of angles are performed under L=3. }
  \label{tab:oriented-decoder}
\end{table}

\subsection{Visualization}
\subsubsection{Quantitative Results}
We qualitatively compare our model with the baseline to provide a comprehensive understanding. As shown in \cref{fig:scale}, our model excels at identifying targets across various scales accurately, even within noisy backgrounds and at different angles. In contrast, the baseline model exhibits shortcomings such as missing parts and noticeable shifts in predicted masks.

\begin{figure}
  \centering
  \includegraphics[width=\linewidth]{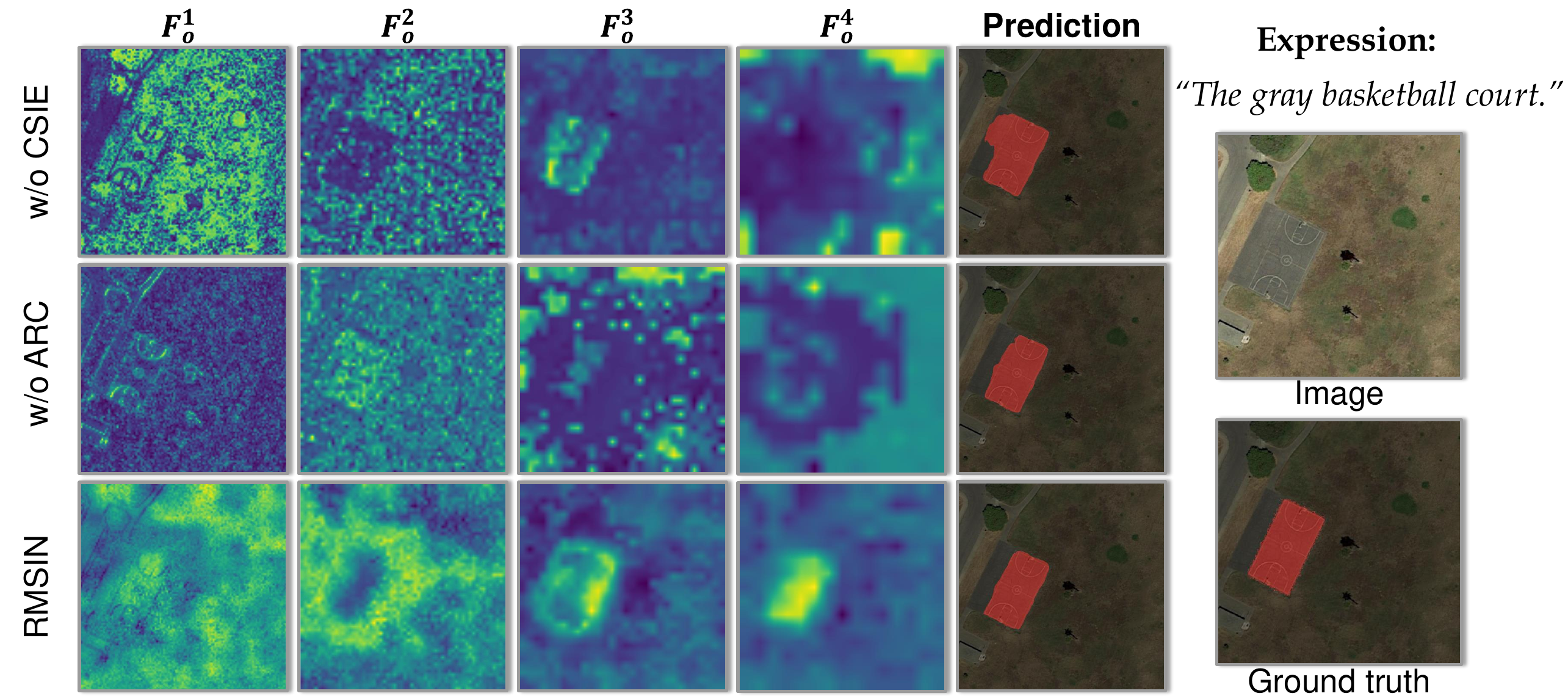}
   \caption{Visualization of predictions and feature maps across stages, where $F_0^{i}$ denotes the feature map for stage $i$. Each row shows the outcomes from progressively adding modules. 
   }
   \label{fig:feature}
   \vspace{-1em}
\end{figure}

\subsubsection{Visualization of Features from Encoder}
In \cref{fig:feature}, we visualize the feature maps from the RMSIN during training under the ablation of ARC and CSIE. It's obvious that RMSIN can accurately capture boundary information with the assistance of scale interaction and rotated convolution. With the scale interaction performed by CSIE and the orientation extraction performed by ARC, RMSIN can focus more keenly on the referred targets. Compared with the first row, CSIE provides more accurate semantics in the deeper layer, and ARC supplies the space prior, which is important for rotated object segmentation.

These qualitative comparisons underscore the efficacy of our approach in addressing challenges related to scale variations and orientation robustness, affirming its capabilities in diverse scenarios.

%% file: sec/6_conclusion.tex
\section{Conclusion}

In this paper, we propose RMSIN, a novel method adept at navigating the complex scales and orientations prevalent in RRSIS. By integrating the IIM and CIM, RMSIN capably addresses the wide range of spatial scales encountered in aerial imagery. Additionally, the implementation of the ARC offers a solid strategy for tackling various orientations. The construction of our expansive RRSIS-D dataset, featuring 17,402 image-caption-mask triplets provides an unparalleled resource in terms of scale and variety for rigorous evaluation. The comprehensive validation on the RRSIS-D dataset not only underscores RMSIN's superior performance but also establishes a new benchmark for future research in this domain.

\section*{Acknowledgement}
This work was supported by National Key R\&D Program of China (No.2023YFB4502804), the National Science Fund for Distinguished Young Scholars (No.62025603), the National Natural Science Foundation of China (No. U21B2037, No. U22B2051, No. 62072389), the National Natural Science Fund for Young Scholars of China (No. 62302411), China Postdoctoral Science Foundation (No. 2023M732948), the Natural Science Foundation of Fujian Province of China (No.2021J01002,  No.2022J06001), and partially sponsored by CCF-NetEase ThunderFire Innovation Research Funding (NO. CCF-Netease 202301).